\documentclass{article}

\usepackage[preprint]{corl_2026} 
\usepackage{graphicx}
\usepackage{amsmath}

\usepackage{amsmath}                
\usepackage{amssymb}                
\usepackage{amsthm}                 
\usepackage{mathtools}              
\usepackage{bm}
\usepackage{graphicx}               
\usepackage{subcaption}             
\usepackage{float}                  
\usepackage[dvipsnames]{xcolor}     
\usepackage{booktabs}               
\usepackage{multirow}               
 
\usepackage{algorithm}
\usepackage{algorithmic}         
\usepackage{enumitem}               
 
\usepackage[nameinlink]{cleveref}   

\title{RoboShape: Information-Theoretic Point Cloud Representations for Privacy-Aware Robot Perception}

\author{
Oguzhan Baser\thanks{The first two co-authors contributed equally.}\\
The University of Texas at Austin \\
\And
H. Mirac Sozen$^*$\\
Bogazici University  \\
\And
Kaan Kale \\
Georgia Institute of Technology \\
\And
Sandeep Chinchali \\
The University of Texas at Austin\\
\And
Sriram Vishwanath \\
Georgia Institute of Technology\\
}

\begin{document}
\maketitle

\begin{abstract}
    With the increased adoption of robotic agents operating in human environments by scanning and sharing 3D representations (e.g., for fleet learning, cloud-based planning, or collaborative mapping), collected point clouds reveal not just the objects in a scene but also sensitive spatial context, such as room function or information that occupants never consented to disclose. Traditional point cloud encoders offer no principled control over this: either all is preserved, or none. Hence, we introduce RoboShape, an information theory guided compression head following the frozen {\tt Sonata} encoder. We project voxel-level embeddings using the Donsker-Varadhan formulation of mutual information (MI). Specifically, we maximize the MI between embeddings and object-level understanding while minimizing it for private attributes. RoboShape leads to 87.5\% smaller embeddings that retain 98.7\% of object classification utility while collapsing sensitive attribute predictions by 39.3\% across the three real-world indoor LiDAR datasets. Its privacy-preserving embeddings are cheaper to transmit over the network or to train a model for any downstream tasks. We release the RoboShape codebase to give the robotics community a practical, encoder-agnostic tool for building perception pipelines that are compact, privacy-aware, and deployment-ready.
\end{abstract}


\section{Introduction}
\label{sec:intro}

3D scene understanding is fundamental to deploying robots in human environments~\cite{survey}. Advances in point cloud encoders, from PointNet ~\cite{qi2017pointnet,qi2017pointnet++} to self-supervised models such as {\tt Sonata}~\cite{sonata}, now enable robots to extract rich spatial representations from depth sensors and LiDAR. These representations power capabilities ranging from semantic object manipulation~\cite{manipulation} and autonomous navigation in cluttered spaces~\cite{navigation} to collaborative mapping and fleet-wide learning across multi-robot systems~\cite{oa1, oa2, oa3}. These new practical capabilities allowed robots to move from controlled laboratories into homes, hospitals, and workplaces, sharing compact 3D representations with cloud services and across agents is becoming central to scalable and capable deployment~\cite{deployment}. Yet the richness that makes modern encoders effective also makes them revealing: a 3D indoor scan of a building encodes not only the furniture a robot needs to interact with but also the function of each room (e.g., bedroom, surveillance room, or IDF room) exposing sensitive spatial context that occupants never intended to disclose. This tension between perceptual utility and spatial privacy remains largely unexamined for 3D point cloud representations in robotics.

\begin{figure*}[t]
    \centering
    \includegraphics[width=0.75\textwidth]{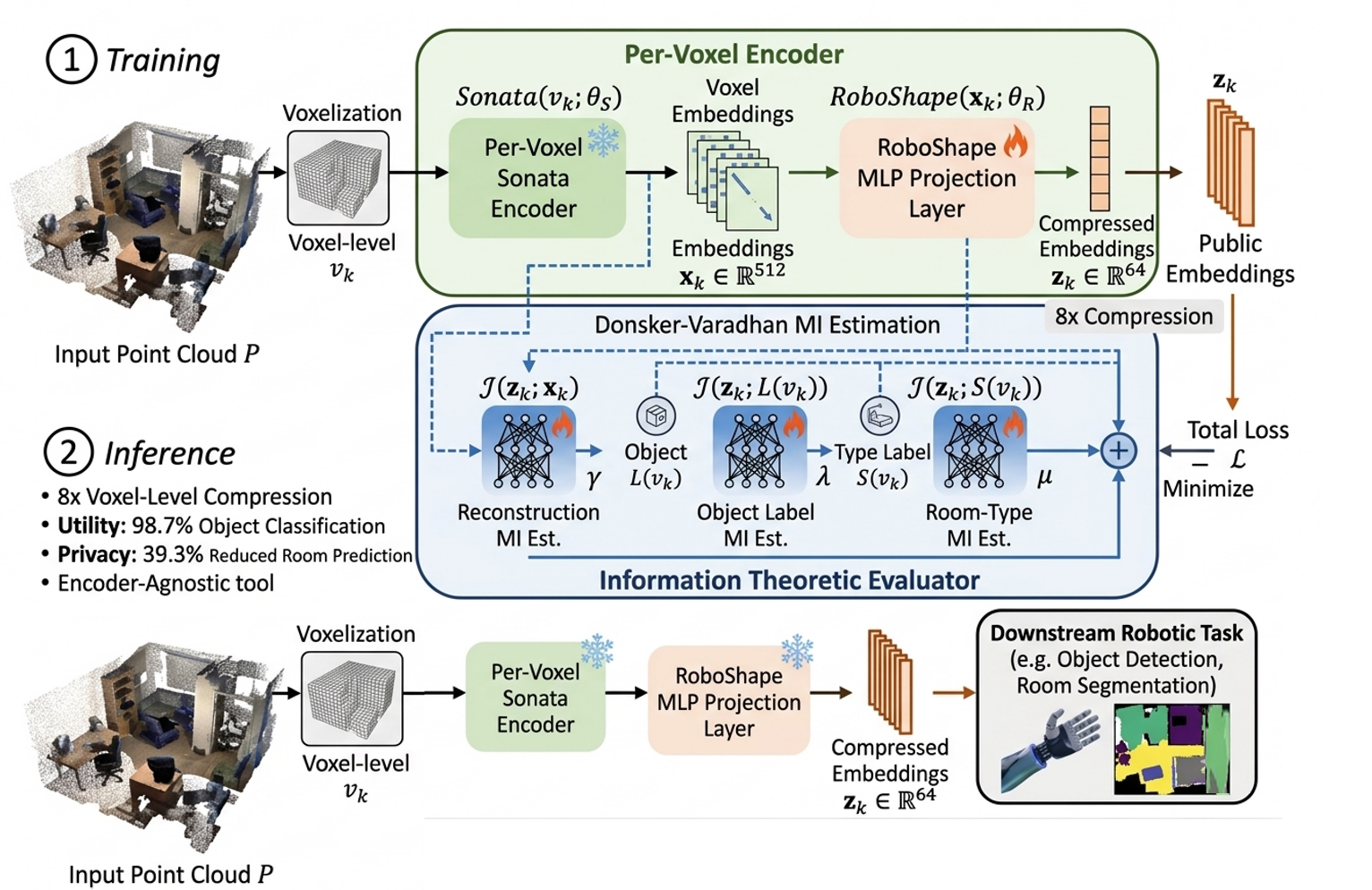}
    \caption{\small \textbf{How does {\tt RoboShape} learn privacy-aware 3D representations?} Our framework first voxelizes a raw point cloud and extracts per-voxel embeddings $\mathbf{x}_k \in \mathbb{R}^{512}$ using a frozen, pre-trained {\tt Sonata} encoder \textit{(green)}. These embeddings, which encode both task-relevant object features and sensitive room-type information, are passed through the {\tt RoboShape} projection layer $\phi_\Theta$ \textit{(orange)} to produce compressed embeddings $\mathbf{z}_k \in \mathbb{R}^{64}$. During training, an information-theoretic evaluator \textit{(blue)} computes three Donsker-Varadhan MI estimates: between $\mathbf{z}_k$ and the original embedding, the object label $L(v_k)$, and the room-type label $S(v_k)$. These serve as differentiable loss terms to maximize task utility while minimizing privacy leakage. At inference, only the frozen encoder and trained projection layer are retained; the MI estimator is discarded. The resulting pipeline compresses each voxel embedding by $87.5\%$. These embeddings can be fed to any downstream tasks.}
    \label{fig:architecture}
\end{figure*}

Despite this growing concern, existing 3D encoders maximize general-purpose feature quality with no mechanism to separate task-relevant attributes from sensitive ones~\cite{encoder_survey}. When a robot shares its scene embedding, all encoded information travels together: object geometry, room layout, and any functional cues the encoder has implicitly captured. Naive mitigation strategies fall short. Adding differential privacy like noise~\cite{dwork2006} degrades the entire representation indiscriminately, suppressing useful structure alongside sensitive attributes. Shuffling projections \cite{neuracrpyt} compress for more privacy without control over which information survives during the dimensionality reduction. Adversarial methods~\cite{adversarial} can learn to suppress specific attributes but lack a measurable, information-theoretic objective that quantifies how much sensitive content actually persists in the output. In adjacent domains, approaches grounded in mutual information (MI) have shown that representations can be precisely shaped to retain targeted features while filtering sensitive ones, with demonstrated success in text~\cite{texshape} and speech~\cite{wavshape}. However, extending MI-based shaping to 3D point clouds introduces distinct challenges: sparse and irregular data structures, variable-density voxel grids, and high-dimensional per-voxel embeddings that make standard MI estimation intractable.

\textbf{Our key observation} is that current approaches treat information suppression as a post-hoc correction rather than a constrained design objective of the representation itself. \textbf{Our technical insight} is that when the scene is decomposed into voxel-level embeddings, operating with those embeddings to estimate the guiding information theoretic constraints is much more practical for robotic deployments. Each voxel representation from a frozen encoder can serve as an independent unit for MI estimation. This decomposition sidesteps the difficulty of estimating information quantities over variable-size point clouds, reducing it to well-conditioned optimization over fixed-dimensional voxel features. Building on this, we introduce {\tt RoboShape} (Fig.~\ref{fig:architecture}), a lightweight projection layer on top of the frozen {\tt Sonata} encoder~\cite{sonata} that reshapes voxel embeddings by maximizing MI with task-relevant object labels while minimizing MI with sensitive room-type attributes. The resulting representation is a compressed, privacy-shaped drop-in replacement for the original encoder output in any downstream robotic perception pipeline. \textbf{Our contributions} can be summarized as follows:\
\begin{enumerate}[leftmargin=*, itemsep=2pt]
    \item We introduce {\tt RoboShape}, the first \textbf{MI-based framework for 3D point cloud data}, extending information-theoretic representation shaping to \textit{embodied robotic perception}.
    \item We propose \textbf{a voxel-level information decomposition for tractable Donsker-Varadhan MI estimation} over sparse and irregular point cloud structures. This way, we \textit{overcome the key technical barrier} to applying information-theoretic methods to 3D scene representations.
    \item  We \textbf{systematically study} the privacy leakage of traditional methods across three real-world datasets: {\tt ScanNet}, {\tt Matterport3D}, and {\tt ARKitScenes}. {\tt RoboShape} achieves \textbf{8$\times$ compression} while retaining \textbf{98.7\%} of object classification utility and reducing private attribute leakage by \textbf{39.3\%}, \textit{consistently outperforms the baselines across both utility and privacy metrics}.
    \item We publicly release our \textbf{full codebase}
    , pre-trained models, and evaluation pipeline to support reproducibility and facilitate future research in privacy-aware robotic perception.
\end{enumerate}

\section{Related Work}
\label{sec:related}
 
We combine three areas: LiDAR representations, information-theory, and privacy-aware perception.
 
Point cloud encoders have evolved from {\tt PointNet}~\cite{qi2017pointnet, qi2017pointnet++} to transformer-based models such as {\tt PointTransformer}~\cite{zhao2021point} and self-supervised strategies including {\tt Sonata}~\cite{sonata} and {\tt Point-MAE}~\cite{pang2022masked}. These produce strong 3D features for manipulation~\cite{manipulation} and navigation~\cite{navigation}, but treat all encoded information equally. No existing encoder controls what its representation reveals. {\tt RoboShape} is complementary. It operates atop any frozen backbone and selectively reshapes its output.
 
Neural MI estimation via the Donsker-Varadhan bound~\cite{mine} enabled differentiable control over representation content. {\tt InfoShape}~\cite{infoshape} demonstrated MI-guided shaping for images, while {\tt TexShape}~\cite{texshape} and {\tt WavShape}~\cite{wavshape} extended the paradigm to text and speech. The closest works to ours are {\tt TexShape} and {\tt WavShape}. They optimize fixed-dimensional sentence or audio-window embeddings. In contrast, {\tt RoboShape} operates on variable-count voxel features from sparse, irregular 3D structures and requires different batching and estimation strategies for tractable MI optimization.
 
Prior work addresses privacy through differential noise~\cite{dwork2006}, federated multi-robot learning~\cite{liu2020fedvision}, and visual anonymization~\cite{pittaluga2019revealing}. These are complementary but rely on heuristics that degrade utility indiscriminately and cannot quantify residual sensitive information. {\tt RoboShape} brings principled MI-based shaping to 3D point clouds, unifying compression, task utility, and privacy suppression under a single measurable objective. This capability is absent from prior work in robotic perception.

\section{Problem Formulation}
\label{sec:problem}

Consider a robot equipped with a sensor that captures a 3D point cloud $\mathcal{P} = \{p_1, \ldots, p_N\}$, where each point $p_i \in \mathbb{R}^3$ lies in a scanned indoor environment, as in Fig.~\ref{fig:architecture}. The robot (or a fleet of robots) must share compact representations of these scans with a cloud server or with other agents for downstream tasks such as object recognition, semantic mapping, or manipulation planning. The central question we address is: \emph{how can a robot compress its 3D scene representation to preserve task-relevant object information while provably suppressing sensitive spatial attributes?}

Following standard practice in 3D scene understanding~\cite{sonata, choy2019minkowski}, we group the point cloud $\mathcal{P}$ into a set of $K$ voxels $\mathcal{V} = \{v_1, \ldots, v_K\}$. Each voxel $v_k$ aggregates the points falling within its spatial cell. A frozen, pre-trained 3D encoder $\psi$, in our case {\tt Sonata}~\cite{sonata}, maps each voxel to $d$-dimensions:
\begin{equation}
    \mathbf{x}_k = \psi(v_k) \in \mathbb{R}^{d}, \quad k = 1, \ldots, K,
    \label{eq:voxel_embed}
\end{equation}
where $d = 512$ for {\tt Sonata}. These per-voxel embeddings form the input to {\tt RoboShape}.

Each voxel inherits two types of annotations. We denote $L(v_k)$ as the \emph{task-relevant} (public) label, corresponding to the object or furniture class the voxel belongs to (e.g., chair, desk, shelf). We denote $S(v_k)$ as the \emph{sensitive} (private) label, corresponding to the room type of the enclosing scene (e.g., bedroom, office, bathroom). From robotics view, $L(v_k)$ captures the geometric understanding a robot needs for manipulation and navigation, while $S(v_k)$ encodes the spatial function that an occupant may wish to keep private when representations are shared externally.

We seek a lightweight, trainable projection $\phi_\Theta: \mathbb{R}^{d} \to \mathbb{R}^{d'}$ that maps each voxel embedding into a lower-dimensional representation $\mathbf{z}_k = \phi_\Theta(\mathbf{x}_k) \in \mathbb{R}^{d'}$, with $d' \ll d$, such that $\mathbf{z}_k$ retains information about $L(v_k)$ while discarding about $S(v_k)$. For notational convenience, we write $L(\mathbf{x})$ and $S(\mathbf{x})$ as shorthand for $L(v_k)$ and $S(v_k)$ when the voxel identity is clear. We formalize this as:
\begin{equation}
    \max_\Theta \;\; \gamma \, I\bigl(\phi_\Theta(\mathbf{x}); \mathbf{x}\bigr) + \lambda \, I\bigl(\phi_\Theta(\mathbf{x}); L(\mathbf{x})\bigr) - \mu \, I\bigl(\phi_\Theta(\mathbf{x}); S(\mathbf{x})\bigr),
    \label{eq:objective}
\end{equation}
where $I(\cdot\,;\cdot)$ denotes mutual information, and $\gamma, \lambda, \mu \geq 0$ are hyperparameters that balance three competing goals: (i) \emph{information-preserving compression}, which ensures that the compressed embedding retains as much of the original representation as possible; (ii) \emph{task utility}, which steers the embedding toward features predictive of object-level labels; and (iii) \emph{privacy}, which suppresses encoding of room-type information. The ratio $d/d'$ controls the compression rate.

This formulation differs from standard 3D representation learning in two ways. First, it introduces an explicit privacy-utility trade-off into the point cloud encoding pipeline, a concern absent from existing encoders~\cite{qi2017pointnet,qi2017pointnet++,sonata}, which maximize general-purpose feature quality without controlling what the representation reveals. Second, the per-voxel structure distinguishes it from prior information-shaping work on text~\cite{texshape} and speech~\cite{wavshape}, where fixed-dimensional embeddings make MI estimation straightforward. In 3D scenes, $K$ varies across scans, and MI must be estimated over spatially irregular structures. This makes the problem  practically important and technically challenging.

\section{Information-Theoretic Approach}
\label{sec:method}

We now describe how {\tt RoboShape} solves the optimization in Eq.~\ref{eq:objective}. The core idea is to treat each per-voxel embedding as an independent sample for MI estimation. This makes the information-theoretic optimization tractable over sparse, irregular 3D point clouds.

\textbf{\underline{Mutual Information Estimation via Donsker-Varadhan:}} Computing the MI between high-dimensional random variables requires access to their joint and marginal distributions, which are generally intractable. Following~\cite{mine, donsker1975}, we adopt the Donsker-Varadhan representation of the Kullback-Leibler divergence, which expresses MI as:
\begin{equation}
    I(\mathbf{A}; \mathbf{B}) = \sup_{F: \Omega \to \mathbb{R}} \; \mathbb{E}_{P_{\mathbf{A},\mathbf{B}}}[F(\mathbf{A}, \mathbf{B})] - \log \mathbb{E}_{P_\mathbf{A} P_\mathbf{B}}[e^{F(\mathbf{A}, \mathbf{B})}],
    \label{eq:dv}
\end{equation}

where $\Omega$ is the joint sample space of $(\mathbf{A}, \mathbf{B})$, and
the supremum is over all functions for which both expectations
are finite. $F$ is parameterized as a neural network $F_\omega$ and trained via SGD~\cite{robbins1951}, with expectations replaced by empirical averages over mini-batches sampled from $P_{\mathbf{A},\mathbf{B}}$ and $P_\mathbf{A} P_\mathbf{B}$.

\textbf{\underline{RoboShape Architecture:}} It consists of three main components as illustrated in Fig.~\ref{fig:architecture}.

\underline{\textit{3D Encoder (Frozen)}:}
A pre-trained {\tt Sonata} encoder processes each voxelized scene and produces per-voxel embeddings $\mathbf{x}_k \in \mathbb{R}^{512}$. The encoder weights are frozen throughout training, ensuring that {\tt RoboShape} operates as a lightweight post-processing module that can sit atop any 3D backbone.

\underline{\textit{Projection Layer (Trainable):}} A shallow neural network $\phi_\Theta$ maps each voxel embedding from $\mathbb{R}^{512}$ to $\mathbb{R}^{64}$. This layer is the only trainable component. Its parameters $\Theta$ are optimized to satisfy the  objective in Eq.~\ref{eq:objective}. At inference time, $\phi_\Theta$ compresses each voxel embedding in parallel, producing representations that are $8\times$ more compact while encoding controlled information content.

\underline{\textit{IT Evaluator (Training Only):}} The MI estimator networks guide the differentiable loss functions during training. Each estimator $F_{\omega_i}$ takes as input the concatenation of the compressed embedding $\mathbf{z}_k$ and the corresponding quantity (original embedding $\mathbf{x}_k$, task label $L(v_k)$, or sensitive label $S(v_k)$) and outputs a scalar used to compute the DV bound. We discard these networks after training.

The key design choice that enables tractable MI estimation over 3D point clouds is treating per-voxel embeddings as independent samples. For a scene with $K$ voxels, we form mini-batches by pooling voxel embeddings across multiple scenes, yielding a batch of fixed-dimensional vectors $\{\mathbf{z}_k\}$ regardless of the varying number of voxels per scene. Joint samples are constructed by pairing $(\mathbf{z}_k, L(v_k))$ from the same voxel; marginal samples are formed by shuffling the label assignments within the batch. This adapts the {\tt RoboShape} to the spatially varying structure of 3D scenes.

Training alternates between two levels of optimization. \textit{(I) Iteration-level (MI estimator update):} Within each epoch, the DV estimator networks $F_{\omega_i}$ are updated for several iterations to refine MI estimates. Mini-batches of voxel embeddings and labels are sampled from the training set, and the estimators are optimized via SGD to maximize the DV bound in Eq.~\ref{eq:dv}. \textit{(II) Epoch-level (Encoder update):} After MI estimates converge, the projection layer parameters $\Theta$ are updated to maximize the combined objective in Eq.~\ref{eq:objective}. As training progresses, fewer MI estimation iterations are needed per epoch, accelerating convergence. After training, the MI evaluator networks are discarded. The frozen {\tt Sonata} encoder paired with the trained projection layer $\phi_\Theta$ forms the complete {\tt RoboShape}.

\section{Experimental Results}
\label{sec:experiments}
\begin{figure}[t]
    \centering
    \includegraphics[width=0.7\textwidth]{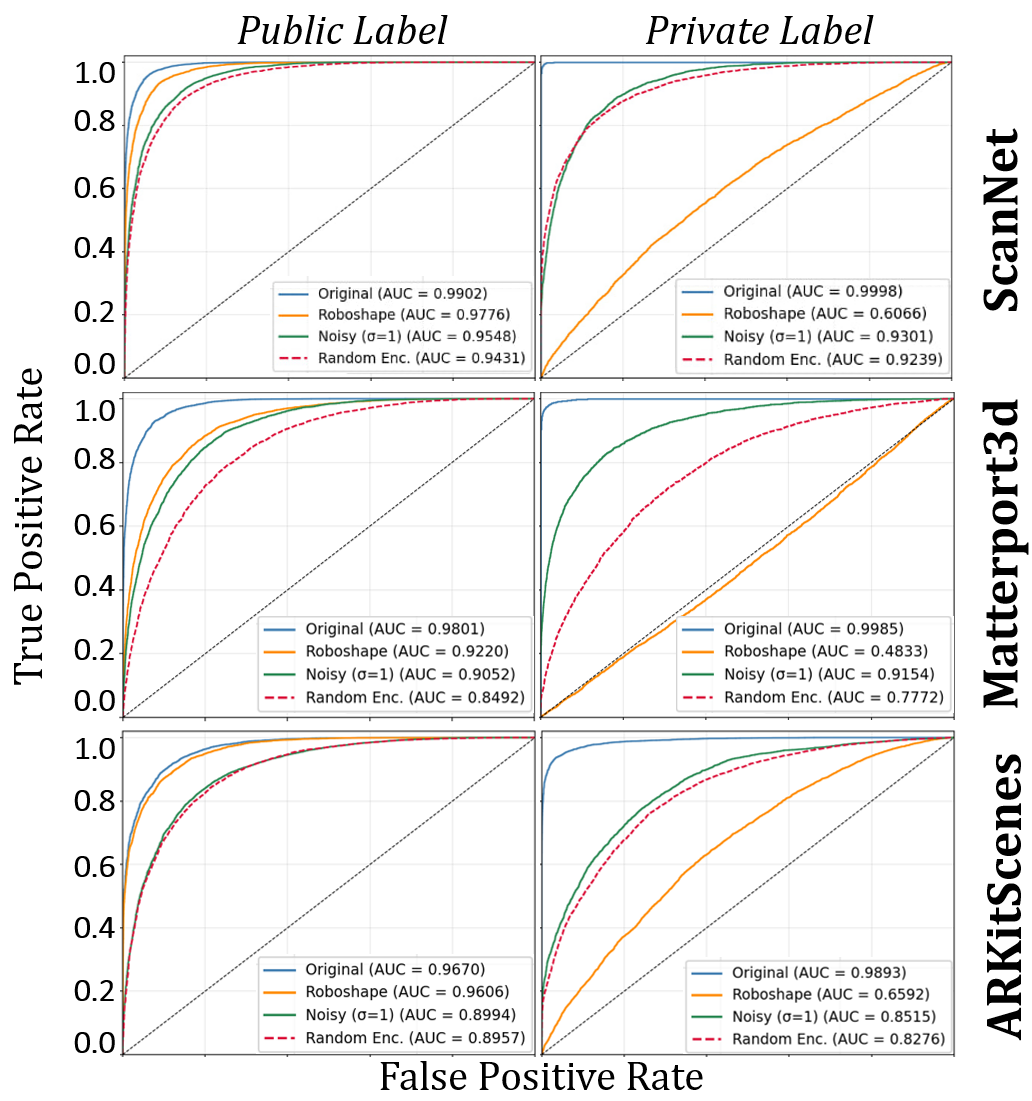}
    \caption{\small \textbf{Can a robot recognize every object in a room without learning what room it is in?} ROC curves and AUROC scores for classifiers trained on four embedding types to predict a task-relevant furniture label (left) and a sensitive room-type label (right). For the task-relevant label, {\tt RoboShape} (orange, AUROC 0.978) nearly matches the original {\tt Sonata} embeddings (blue, AUROC 0.990), retaining 98.7\% of classification utility despite $8\times$ compression. For the sensitive label, {\tt RoboShape} collapses the AUROC from 0.9998 to 0.607 while both the {\tt Noisy} baseline (green, AUROC 0.930) and {\tt Random} (red dashed, AUROC 0.943) leak substantially more private information. {\tt RoboShape} simultaneously achieves high task utility and low sensitive leakage.}
    \label{fig:roc_curves}
\end{figure}

We compare {\tt RoboShape} against \textit{three baselines} that are common for private learned representations.

{\tt Original}: {\tt Sonata} embeddings serves as the upper bound for both utility and privacy leakage.\\
{\tt Noisy}: {\tt Original} is injected Gaussian differential privacy noise by suppressing indiscriminately.\\
{\tt Random Encoder}: A randomly initialized projection layer (same architecture as {\tt RoboShape}, untrained), which compresses embeddings without any information-theoretic guidance similar to \cite{neuracrpyt}.

We exclude adversarial training as it lacks measurable objectives to quantify residual sensitive data. Then, we evaluate baselines on \textit{three established 3D indoor scene datasets} and report results using \textit{three complementary metrics}: mutual information evolution during training, classification AUROC for downstream probing, and qualitative t-SNE visualizations.We describe the datasets as follows.

\textbf{{\tt ScanNet}~\cite{scannet}:} An RGB-D dataset of 1,513 densely annotated indoor reconstructions across 707 distinct rooms, with per-point semantic labels spanning 20 object categories. Its single-room scale and clean structured-light captures provide a controlled setting to establish core privacy-utility performance, isolating the information-shaping effect from sensor noise and scene complexity.
 
\textbf{{\tt Matterport3D}~\cite{matterport3d}:} A dataset of 90 building-scale RGB-D reconstructions spanning residential, commercial, and institutional buildings with 40 semantic categories. Its multi-room structure and diverse room types (bedrooms, bathrooms, offices, hallways) test whether privacy suppression generalizes across spatial scales and richer room-type distributions than single-room benchmarks.

\textbf{{\tt ARKitScenes}~\cite{dehghan2021arkitscenes}:} The largest publicly available indoor scene understanding dataset, comprising 5,048 RGB-D sequences across 1,661 unique scenes captured with Apple's mobile LiDAR sensor. Unlike the stationary scanning setups of {\tt ScanNet} and {\tt Matterport3D}, {\tt ARKitScenes} reflects the noisy, partial observations a mobile robot would encounter in practice, stress-testing {\tt RoboShape}.

\begin{figure}[t]
    \centering
    \includegraphics[width=\columnwidth]{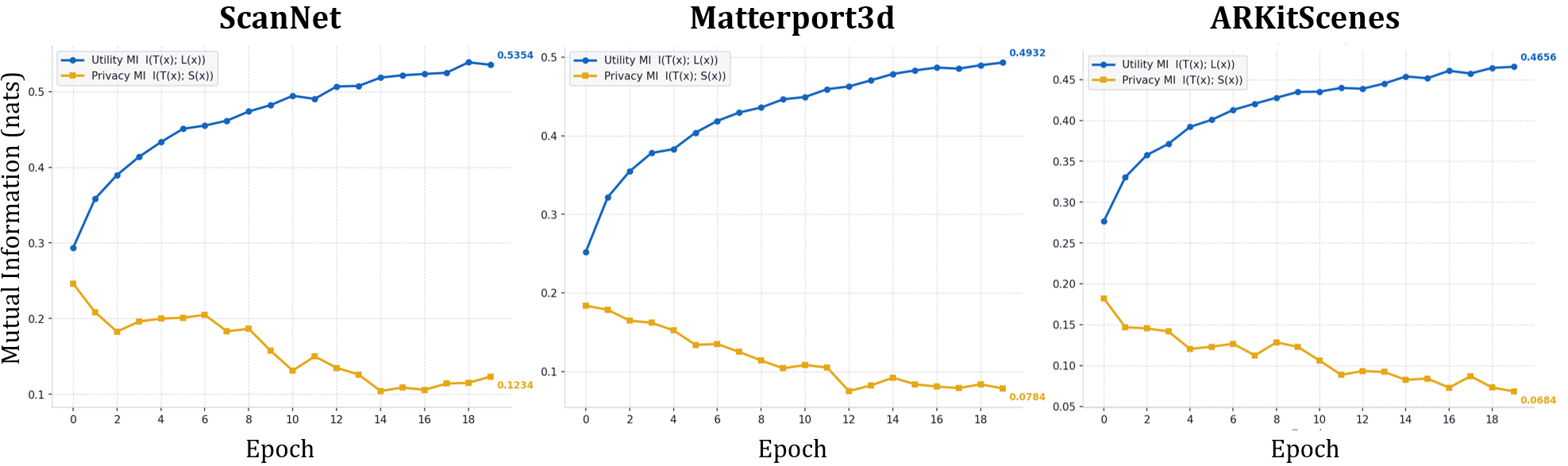}
    \caption{\small \textbf{How does MI loss disentangle task-relevant and sensitive information in point clouds?} Evolution of utility MI $I(\phi_\Theta(\mathbf{x}); L(\mathbf{x}))$ (blue) and privacy MI $I(\phi_\Theta(\mathbf{x}); S(\mathbf{x}))$ (yellow) over 20 training epochs. Utility MI rises steadily from 0.293 to 0.535 nats while privacy MI drops from 0.247 to 0.123 nats, a 50\% reduction. The two curves diverge monotonically without oscillation. The weighted objective in Eq. \ref{eq:objective} successfully steers the projection layer to concentrate object-level structure into the compressed embedding while progressively filtering out room-type information.}
    \label{fig:mi_curves}
\end{figure}

\textit{\textbf{Does Privacy Shaping Preserve What a Robot Needs to Know?}}

To translate MI into actionable metrics, we train lightweight binary classifiers on the compressed embeddings to predict \textit{(a)} a task-relevant furniture label and \textit{(b)} a sensitive room-type label. Fig.~\ref{fig:roc_curves} presents the ROC curves for all four embedding types.

\textbf{Task-relevant label:} {\tt RoboShape} achieves an AUROC of 0.978, compared to 0.990 for the original {\tt Sonata} embeddings, retaining \textbf{98.7\%} of classification utility despite \textbf{87.5\%} compression. Both the {\tt Noisy} baseline (AUROC of 0.955) and the {\tt Random} encoder (AUROC of 0.943) perform notably worse. Our MI-based optimization preserves task structure more effectively than heuristic methods.

\textbf{Sensitive label:} {\tt Original} embeddings yield a near-perfect AUROC of 0.9998 for room-type prediction, confirming that unmodified {\tt Sonata} features heavily encode spatial context. {\tt RoboShape} collapses this by 39.3\% to AUROC of 0.607 while the {\tt Noisy} (AUROC of 0.930) and {\tt Random} (AUROC of 0.924) baselines still leak substantially more private information. It shows our work produces representations that recognize objects in a space without revealing what kind of space it is.

\textbf{\textit{Can Utility and Privacy Be Disentangled Without Training Instability?}}

Fig.~\ref{fig:mi_curves} tracks the evolution of utility MI $I(\phi_\Theta(\mathbf{x}); L(\mathbf{x}))$ and privacy MI $I(\phi_\Theta(\mathbf{x}); S(\mathbf{x}))$ over 20 training epochs. Two trends emerge clearly. First, utility MI rises steadily from 0.293 to 0.535 nats, confirming that the projection layer learns to concentrate task-relevant structure into the compressed embedding. Second, privacy MI drops from 0.247 to 0.123 nats (a 50\% reduction) showing that room-type information is progressively filtered out. The two curves diverge monotonically, demonstrating that the weighted objective in Eq.~\ref{eq:objective} successfully disentangles the competing information streams without oscillation or instability.

\textbf{\textit{What Happens to the Embedding Geometry After Information Shaping?}}

\begin{figure}[t]
    \centering
    \includegraphics[width=\linewidth]{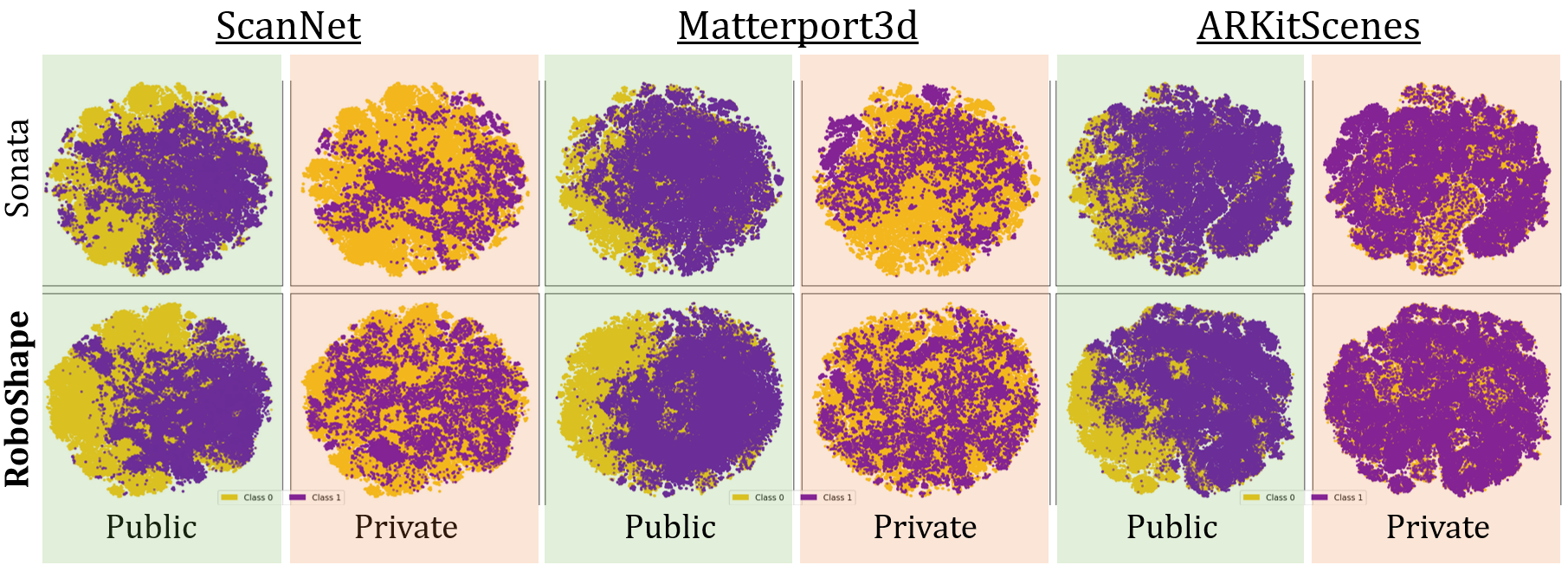}
    \caption{\small \textbf{What happens to the geometry of the embedding space after information shaping?} t-SNE visualizations of per-voxel embeddings colored by furniture type (left) and room type (right), before (top) and after (bottom) applying {\tt RoboShape}. In the original {\tt Sonata} embeddings, voxels form distinct clusters under both coloring schemes, confirming that the pre-trained encoder encodes task-relevant and sensitive attributes alike. After {\tt RoboShape}, furniture-based clusters remain clearly separable, consistent with the 98.7\% AUROC retention in Fig.~\ref{fig:roc_curves}, while room-type clusters dissolve into an overlapping distribution, qualitatively confirming Fig.~\ref{fig:mi_curves}. The projection layer reshapes the embedding geometry to preserve task-relevant structure while erasing the sensitive spatial signal.}
    \label{fig:tsne}
\end{figure}

Fig.~\ref{fig:tsne} visualizes the embedding space before and after {\tt RoboShape} using t-SNE~\cite{tsne}. In {\tt Original} {\tt Sonata} embeddings, voxels cluster distinctly by both furniture type and room type. After {\tt RoboShape}, the furniture-based clusters remain intact while room-type clusters dissolve into an overlapping distribution, consistent with the MI and AUROC results. This confirms that the information-theoretic objective reshapes the geometry of the embedding space in the intended direction: the representation retains task-relevant separability while smoothing out the sensitive attribute.

\textbf{\textit{Does Information Shaping Survive Closed-Loop Robotic Control?}}
 
Classification AUROC confirms that {\tt RoboShape} embeddings suppress room-type information statistically but a robot acts on its representations in closed loop, where even residual information can be exploited by a learned policy. To test whether MI-based shaping holds under embodied control, we train PPO policies in NVIDIA Isaac Gym~\cite{makoviychuk2021isaac} on 52 parallel ScanNet-derived environments (40 training, 12 held-out test), each populated with voxelized 3D scenes and a point-mass agent that observes per-voxel embeddings. Training and test environments contain balanced mixtures of bedroom and non-bedroom scenes to prevent the policy from exploiting distributional shortcuts.
 
\begin{figure}[t]
    \centering
    \includegraphics[width=\linewidth]{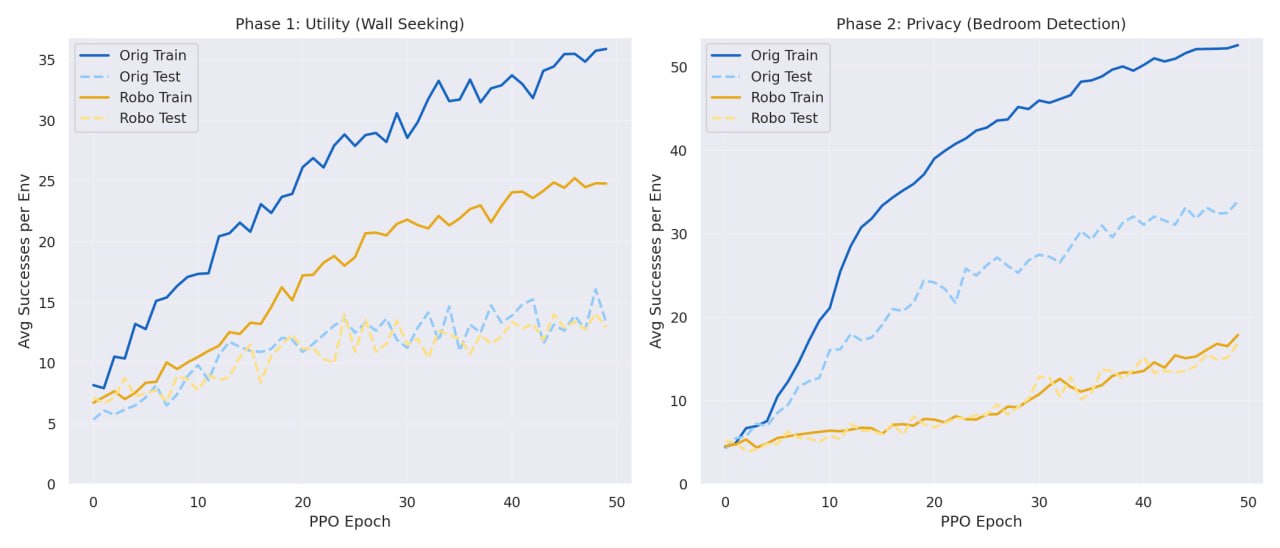}
    \caption{\small \textbf{Does a PPO policy exploit private information when acting on {\tt RoboShape} embeddings?} Average successes per environment over 50 PPO epochs across 52 Isaac Gym environments. \textit{Left (Utility):} wall-seeking navigation, where {\tt RoboShape} (orange) retains meaningful task performance relative to original {\tt Sonata} embeddings (blue). \textit{Right (Privacy):} bedroom-conditional navigation, where original embeddings enable rapid room-type exploitation ($\sim$52 train successes) while {\tt RoboShape} reduces this by 63\% to $\sim$19. This confirms that MI-based shaping transfers from static classification to closed-loop embodied control.}
    \label{fig:isaacgym}
\end{figure}

We evaluate two phases. In \textbf{Phase~1 (Utility)}, the agent must navigate to walls (a task-relevant target) using either {\tt Original} {\tt Sonata} embeddings ($\mathbb{R}^{512}$) or {\tt RoboShape} embeddings ($\mathbb{R}^{64}$). Fig.~\ref{fig:isaacgym} (left) shows that {\tt RoboShape} policies reach $\sim$69\% of the original's training success rate. It shows out compressed features retain sufficient geometric structure for goal-directed navigation. In \textbf{Phase~2 (Privacy)}, the agent must navigate conditionally on room type specifically, moving to one of two distinct spatial zones depending on whether the scene is a bedroom or a non-bedroom. Here, original embeddings enable the policy to rapidly exploit room-type information (reaching $\sim$52 successes), while {\tt RoboShape} embeddings suppress this signal to $\sim$19 successes, a 63\% reduction (Fig.~\ref{fig:isaacgym}, right). The gap between the two curves is the embodied manifestation of the MI and AUROC results: a robot acting on {\tt RoboShape} features can find walls but cannot determine whether it is in a bedroom, even when a PPO policy is explicitly optimized to do so.

\subsection{Limitations}
\label{sec:limitations}

Our framework inherits several assumptions worth noting. First, the voxel-level independence assumption treats each voxel embedding as an i.i.d.\ sample for MI estimation, which may not hold for spatially correlated voxels within the same object. Second, our current evaluation uses scene-level room-type labels; finer-grained privacy attributes (e.g., per-object ownership) would require richer annotations. Third, while the Isaac Gym experiments demonstrate embodied
performance on navigation, we have not yet evaluated {\tt RoboShape} on contact-rich manipulation or long-horizon planning tasks where
representation quality manifests differently. We address these directions in future work.

\section{Conclusion and Future Work}
\label{sec:conclusion}

We introduce {\tt RoboShape}, an information-theoretic framework for learning compressed, privacy-aware 3D point cloud representations for robot perception. By training a lightweight projection layer atop a frozen Sonata encoder using mutual information optimization via the Donsker-Varadhan formulation,  {\tt RoboShape} achieves \textbf{87.5\%} embedding compression while retaining \textbf{98.7\%} of object classification utility and reducing room-type leakage from near-perfect to near-chance levels (AUROC from 0.9998 to 0.607). Across {\tt ScanNet}, {\tt Matterport3D}, and {\tt ARKitScenes},  we consistently outperform the baselines on both utility retention and privacy suppression. This validates that MI-based feature shaping produces better trade-offs than heuristic alternatives for 3D representations.

We treat per-voxel embeddings from a frozen 3D encoder as independent estimation units for MI optimization. This enables tractable information-theoretic control over what irregular, sparse 3D representations encode. This bridges information-theoretic data shaping into embodied perception, providing the robotics community with a practical, encoder-agnostic tool for building perception pipelines that are simultaneously compact, privacy-aware, and deployment-ready.

Several directions merit further investigation. First, we plan to evaluate  {\tt RoboShape} representations on more downstream embodied tasks (e.g., navigation, manipulation, and semantic goal-reaching) to validate that privacy-shaped embeddings support closed-loop robotic performance. Second, extending the framework to handle multiple sensitive attributes simultaneously (e.g., room type, building identity, and occupant activity patterns) would broaden its applicability to real-world fleet deployments. Finally, exploring nonlinear MI optimization strategies, such as multi-objective reinforcement learning over the $(\gamma, \lambda, \mu)$ trade-off surface, could automate the hyperparameter selection that currently requires manual tuning. Our code and data are publicly available 
for reproducibility and future research.

\bibliography{example}  
\newpage
\appendix

\appendix

\section*{Appendix}
The appendix is organized as follows:
\begin{itemize}[leftmargin=*, itemsep=1pt]
    \item Section~\ref{app:theory} provides the information-theoretic foundations.
    \item Section~\ref{app:results} reports full numerical results across all datasets.
    \item Section~\ref{app:datasets} details dataset distributions and preprocessing.
    \item Section~\ref{app:architecture} specifies architecture and training details.
    \item Section~\ref{app:isaacgym} describes the IsaacGym simulation setup.
    \item Section~\ref{app:codebase} documents the released codebase.
    \item Section~\ref{app:impact} discusses broader impact.
\end{itemize}

\section{Theoretical Foundations}
\label{app:theory}

This section provides the information-theoretic background underlying {\tt RoboShape}. We derive the Donsker-Varadhan MI estimator, analyze its estimation properties, justify the voxel-level independence assumption, and explain why the bilevel optimization in Eq.~2 converges to a meaningful privacy-utility trade-off.

\subsection{From KL Divergence to the DV Bound}

Mutual information between two random variables $\mathbf{A}$ and $\mathbf{B}$ equals the KL divergence between their joint and the product of their marginals:
\begin{equation}
    I(\mathbf{A}; \mathbf{B}) = D_{\mathrm{KL}}(P_{\mathbf{A},\mathbf{B}} \,\|\, P_\mathbf{A} P_\mathbf{B}).
    \label{eq:mi_kl_app}
\end{equation}
The Donsker-Varadhan representation~\cite{donsker1975} provides a variational characterization of KL divergence:
\begin{equation}
    D_{\mathrm{KL}}(P \,\|\, Q) = \sup_{F} \; \mathbb{E}_{P}[F] - \log \mathbb{E}_{Q}[e^{F}],
    \label{eq:dv_kl_app}
\end{equation}
where the supremum is over all measurable functions $F$ for which both expectations are finite. Substituting Eq. \ref{eq:mi_kl_app} into Eq. \ref{eq:dv_kl_app} yields the MI estimator used in the main paper (Eq.~3). In {\tt RoboShape}, we instantiate three copies of this bound---one per term in the objective (Eq.~2)---with $F$ parameterized as a neural network $F_\omega$ and optimized via gradient ascent, yielding a differentiable lower bound $\hat{I}_{\mathrm{DV}} \leq I$.

\subsection{Estimation Error and Finite-Sample Bias}

The DV estimator is consistent (converges to the true MI as the number of samples $n \to \infty$ and the function class capacity grows) but exhibits finite-sample bias~\cite{mine, song2019}. Two sources contribute:

\paragraph{Log-sum-exp bias.} The second term in Eq. \ref{eq:dv_kl_app} involves $\log \mathbb{E}[e^F]$, which is estimated as $\log \frac{1}{n}\sum_{i=1}^{n} e^{F(\mathbf{a}_i, \mathbf{b}_i)}$. By Jensen's inequality, this empirical estimate is biased upward, causing $\hat{I}_{\mathrm{DV}}$ to underestimate the true MI. We mitigate this with large batch sizes and an exponential moving average on the log-sum-exp term, following~\cite{mine}.

\paragraph{Dimensionality dependence.} Song and Ermon~\cite{song2019} show that the sample complexity of MI estimation scales as $\mathcal{O}(e^{I(\mathbf{A};\mathbf{B})})$ in the worst case. Our $8\times$ compression from $d = 512$ to $d' = 64$ directly reduces the embedding dimension over which MI is estimated, improving the estimator's effective sample efficiency.

\subsection{Why Minimizing a Lower Bound on MI Works}

The objective in Eq.~2 maximizes the first two MI terms (compression and utility) while \emph{minimizing} the third (privacy). Since $\hat{I}_{\mathrm{DV}}$ is a lower bound on the true MI, a natural concern is whether minimizing a lower bound actually reduces the true MI.

The key insight is that {\tt RoboShape} uses a bilevel optimization:
\begin{enumerate}[leftmargin=*, itemsep=2pt]
    \item \textbf{Inner loop:} The MI estimator $F_{\omega_3}$ is trained to \emph{maximize} the DV bound for the privacy term, tightening the lower bound toward the true MI.
    \item \textbf{Outer loop:} The projection layer $\phi_\Theta$ is updated to \emph{minimize} this tightened bound.
\end{enumerate}

This creates a minimax dynamic: $F_{\omega_3}$ tries to detect as much room-type information as possible in $\mathbf{z}$, while $\phi_\Theta$ tries to make room-type information undetectable. As training progresses, if $\phi_\Theta$ succeeds in removing room-type information, no function $F_{\omega_3}$ can recover it, and the DV bound converges to a value near zero. This is analogous to adversarial training, but with the critical advantage that the DV bound provides a \emph{scalar MI estimate} at each step, giving a measurable, interpretable privacy metric rather than a binary adversarial outcome.

Formally, at convergence:
\begin{equation}
    \hat{I}_{\mathrm{DV}}(\mathbf{z}; S(\mathbf{x})) \leq I(\mathbf{z}; S(\mathbf{x})) \leq I(\mathbf{x}; S(\mathbf{x})),
    \label{eq:dpi}
\end{equation}
where the second inequality follows from the \emph{data processing inequality} (DPI): since $\mathbf{z} = \phi_\Theta(\mathbf{x})$ is a deterministic function of $\mathbf{x}$, processing can only reduce MI. {\tt RoboShape} exploits this by designing $\phi_\Theta$ to aggressively reduce the right-hand side for the sensitive attribute while preserving it for the task-relevant attribute.

\subsection{Data Processing Inequality and Compression}

The DPI provides a fundamental guarantee for {\tt RoboShape}: for any deterministic mapping $\phi_\Theta$,
\begin{align}
    I(\mathbf{z}; L(\mathbf{x})) &\leq I(\mathbf{x}; L(\mathbf{x})), \label{eq:dpi_utility} \\
    I(\mathbf{z}; S(\mathbf{x})) &\leq I(\mathbf{x}; S(\mathbf{x})). \label{eq:dpi_privacy}
\end{align}

Eq. \ref{eq:dpi_utility} means that compression necessarily discards \emph{some} task-relevant information---the question is how much. {\tt RoboShape}'s objective steers $\phi_\Theta$ to preferentially retain $L(\mathbf{x})$-relevant features.  Eq. \ref{eq:dpi_privacy} means that {\tt RoboShape} can reduce but never increase privacy leakage relative to the original encoder---a desirable monotonicity property.

The fact that {\tt RoboShape} retains 98.7\% of task AUROC while collapsing privacy AUROC by 39.3\% suggests that $L(\mathbf{x})$ and $S(\mathbf{x})$ are encoded in partially separable subspaces of the original Sonata embedding, and the MI-guided projection successfully identifies and preserves the task-relevant subspace while projecting out the privacy-relevant one.

\subsection{Voxel-Level Independence Assumption}

{\tt RoboShape} treats per-voxel embeddings as independent samples for MI estimation. Formally, for a mini-batch $\mathcal{B}$ of voxels pooled across scenes:
\begin{equation}
    \hat{I}(\mathbf{z}; L(\mathbf{x})) \approx \frac{1}{|\mathcal{B}|} \sum_{k \in \mathcal{B}} \hat{I}_{\mathrm{DV}}(\mathbf{z}_k; L(v_k)).
\end{equation}

This approximation holds well when two conditions are met:

\paragraph{Condition 1: Local receptive field.} The Sonata encoder's hierarchical architecture means that each voxel embedding $\mathbf{x}_k$ depends primarily on points within a local spatial neighborhood. Voxels from distant parts of a scene carry approximately independent information about their semantic labels.

\paragraph{Condition 2: Cross-scene pooling.} By pooling voxels across multiple scenes into each mini-batch, we ensure that the batch contains samples from diverse spatial contexts. This breaks within-scene correlations and makes the i.i.d.\ assumption more realistic at the batch level, even if individual voxels within a single scene exhibit local dependence.

In practice, we observe stable, monotonic MI convergence curves across all three datasets (Fig.~3 in the main paper), with no oscillation or divergence---empirical evidence that the independence assumption is sufficiently accurate for the DV estimator to provide reliable gradient signals.

\subsection{Connection to Rate-Distortion Theory}

The {\tt RoboShape} objective (Eq.~2) can be interpreted through the lens of rate-distortion theory~\cite{cover1999}. The first term $\gamma\, I(\mathbf{z}; \mathbf{x})$ controls the \emph{rate}---how many bits the compressed representation preserves about the original. The second and third terms shape the \emph{distortion} by specifying which bits matter (object-level features) and which should be discarded (room-type features).

Classical rate-distortion theory operates with a single distortion measure, whereas {\tt RoboShape} introduces a \emph{multi-objective} formulation with competing distortion criteria. The hyperparameters $(\gamma, \lambda, \mu)$ trace a Pareto surface in the rate-utility-privacy space. Setting $\gamma = 0$ in our experiments corresponds to being agnostic about the overall rate and focusing entirely on the utility-privacy trade-off, which is the regime most relevant to robotic deployment where downstream task performance and occupant privacy are the primary concerns.

\section{Full Numerical Results}
\label{app:results}

Table~\ref{tab:full_auroc} reports the complete AUROC scores for all four embedding types across all three datasets. Table~\ref{tab:mi_convergence} reports the converged MI values after 20 training epochs. Across all datasets, {\tt RoboShape} consistently achieves the lowest private-label AUROC while maintaining task-relevant performance close to the uncompressed {\tt Original} embeddings.

\begin{table}[h]
\centering
\caption{\textbf{Classification AUROC across all datasets and methods.} Public = task-relevant label (furniture), Private = sensitive label (room type). Best privacy result (lowest private AUROC) per dataset in \textbf{bold}. {\tt RoboShape} is the only method that simultaneously achieves high public AUROC and near-chance private AUROC.}
\label{tab:full_auroc}
\begin{tabular}{lcccccc}
\toprule
& \multicolumn{2}{c}{\textbf{ScanNet}} & \multicolumn{2}{c}{\textbf{Matterport3D}} & \multicolumn{2}{c}{\textbf{ARKitScenes}} \\
\cmidrule(lr){2-3} \cmidrule(lr){4-5} \cmidrule(lr){6-7}
\textbf{Method} & Public & Private & Public & Private & Public & Private \\
\midrule
{\tt Original}               & 0.990 & 1.000 & 0.980 & 0.999 & 0.967 & 0.989 \\
{\tt Noisy} ($\sigma\!=\!1$) & 0.955 & 0.930 & 0.905 & 0.915 & 0.899 & 0.852 \\
{\tt Random Enc.}             & 0.943 & 0.924 & 0.849 & 0.777 & 0.896 & 0.828 \\
\textbf{{\tt RoboShape}}     & 0.978 & \textbf{0.607} & 0.922 & \textbf{0.483} & 0.961 & \textbf{0.659} \\
\bottomrule
\end{tabular}
\end{table}

\begin{table}[h]
\centering
\caption{\textbf{Converged MI values (nats) after 20 training epochs.} Utility MI is maximized (higher is better); privacy MI is minimized (lower is better). Privacy MI drops by 50--60\% across all datasets.}
\label{tab:mi_convergence}
\begin{tabular}{lccc}
\toprule
\textbf{MI Term} & \textbf{ScanNet} & \textbf{Matterport3D} & \textbf{ARKitScenes} \\
\midrule
Utility $I(\mathbf{z}; L(\mathbf{x}))$ & 0.535 & 0.493 & 0.466 \\
Privacy $I(\mathbf{z}; S(\mathbf{x}))$  & 0.123 & 0.078 & 0.068 \\
\bottomrule
\end{tabular}
\end{table}

\section{Dataset Details}
\label{app:datasets}

\begin{figure}
    \centering
    \includegraphics[width=0.85\linewidth]{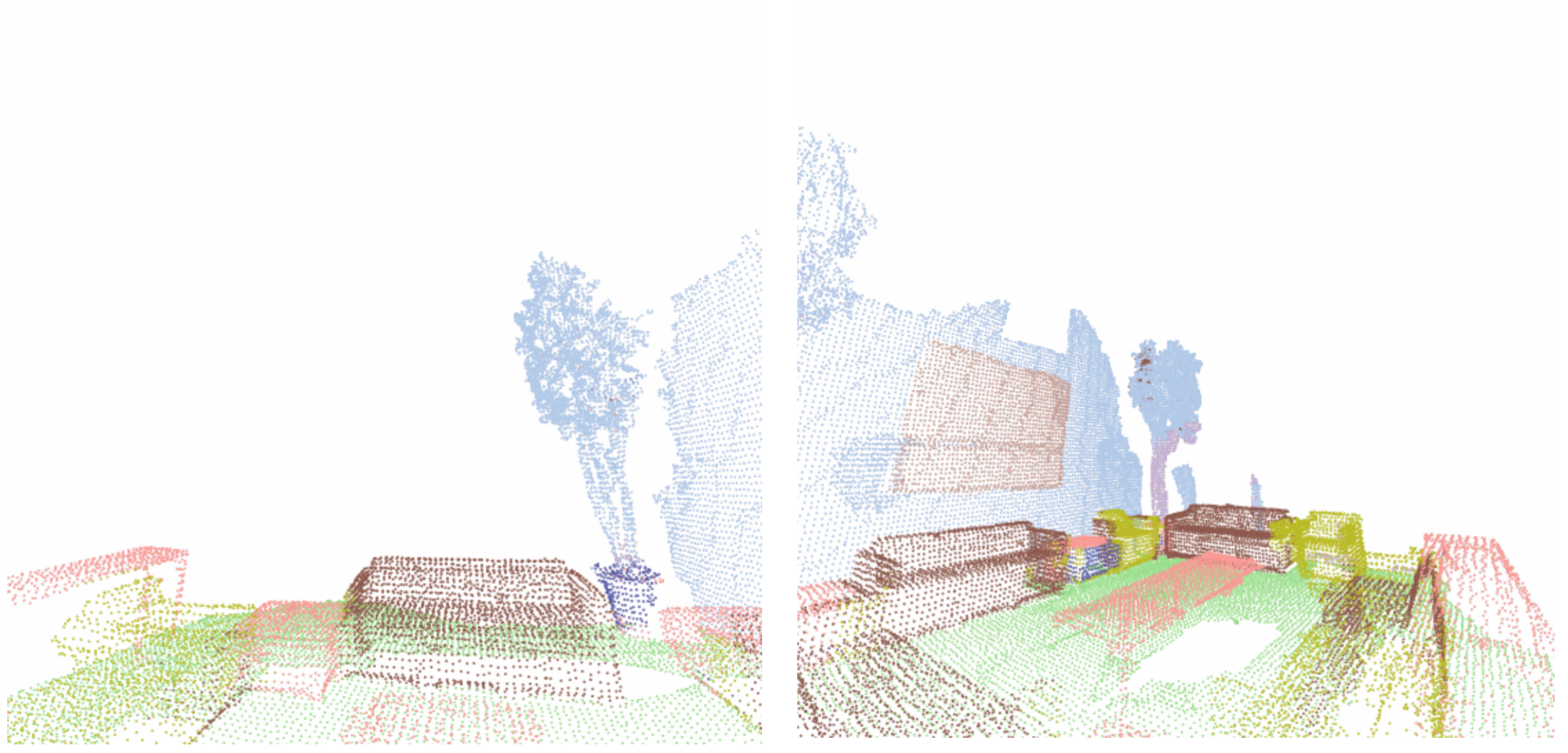}
    \caption{A ScanNet sample colored by Sonata predictions}
\end{figure}

Table~\ref{tab:dataset_stats} summarizes the three datasets. We selected these to cover a progression from controlled (ScanNet) to diverse (Matterport3D) to deployment-realistic (ARKitScenes) conditions.

\begin{table}[h]
\centering
\caption{\textbf{Dataset statistics.} ScanNet provides controlled single-room scans; Matterport3D tests generalization to building-scale environments; ARKitScenes introduces mobile LiDAR noise representative of real robotic deployment.}
\label{tab:dataset_stats}
\begin{tabular}{lccc}
\toprule
& \textbf{ScanNet} & \textbf{Matterport3D} & \textbf{ARKitScenes} \\
\midrule
Total scenes / buildings & 1,513 scans & 90 buildings & 1,661 scenes \\
Train / Val / Test & 1,201 / 312 / --- & 61 / 11 / 18 & 80\% / 10\% / 10\% \\
Object categories & 20 & 40 & 17 \\
Sensor & Structured-light & Matterport Pro & Apple LiDAR \\
Scale & Single-room & Building-scale & Single-room \\
\bottomrule
\end{tabular}
\end{table}

The following analyses focus on ScanNet, our primary benchmark, and illustrate the data characteristics that informed our experimental design.

\begin{figure}[h]
\centering
\includegraphics[width=0.85\textwidth]{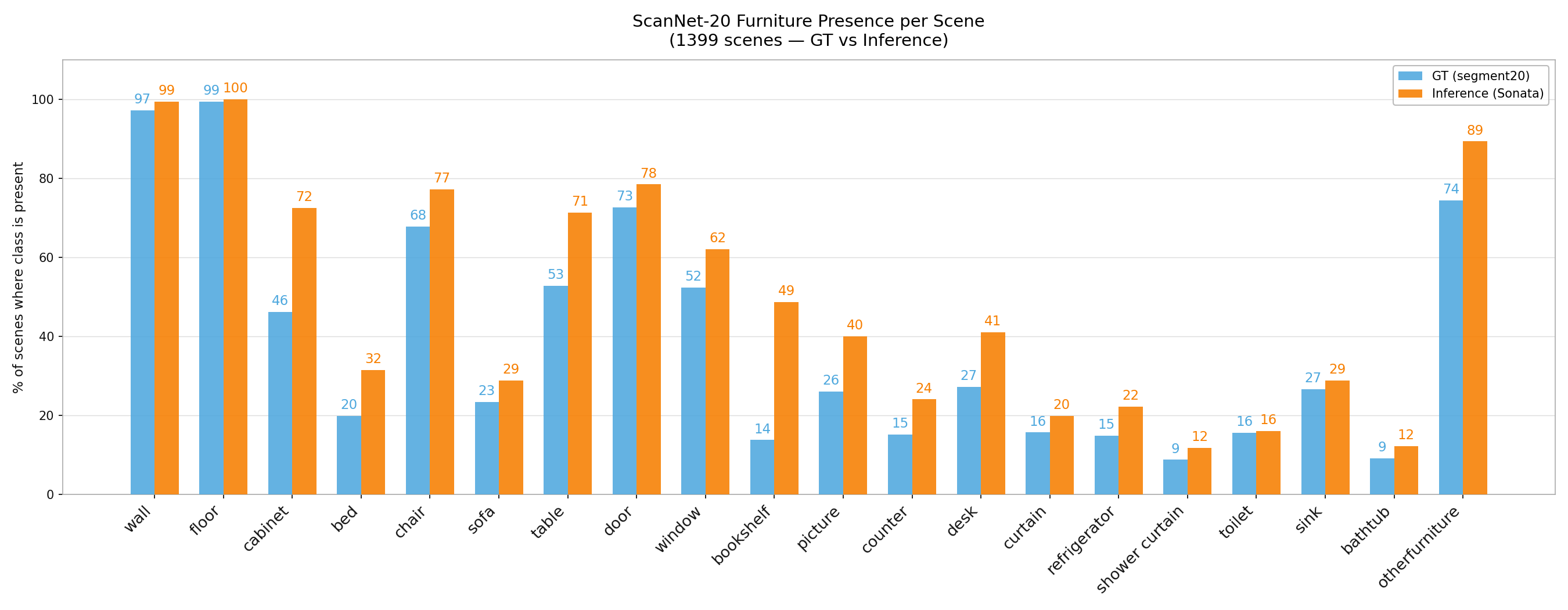}
\caption{Furniture category frequency across ScanNet scenes.}
\label{fig:furniture_by_room}
\end{figure}

\paragraph{Furniture presence across scenes:} Fig.~\ref{fig:furniture_by_room} shows the frequency of each furniture category across ScanNet scenes. Common objects (walls, floors, chairs) appear in nearly every scene, while rarer categories (bookshelves, curtains) occur infrequently. This imbalance motivated our use of binary classification for the utility evaluation.

\begin{figure}[h]
\centering
\includegraphics[width=0.7\textwidth]{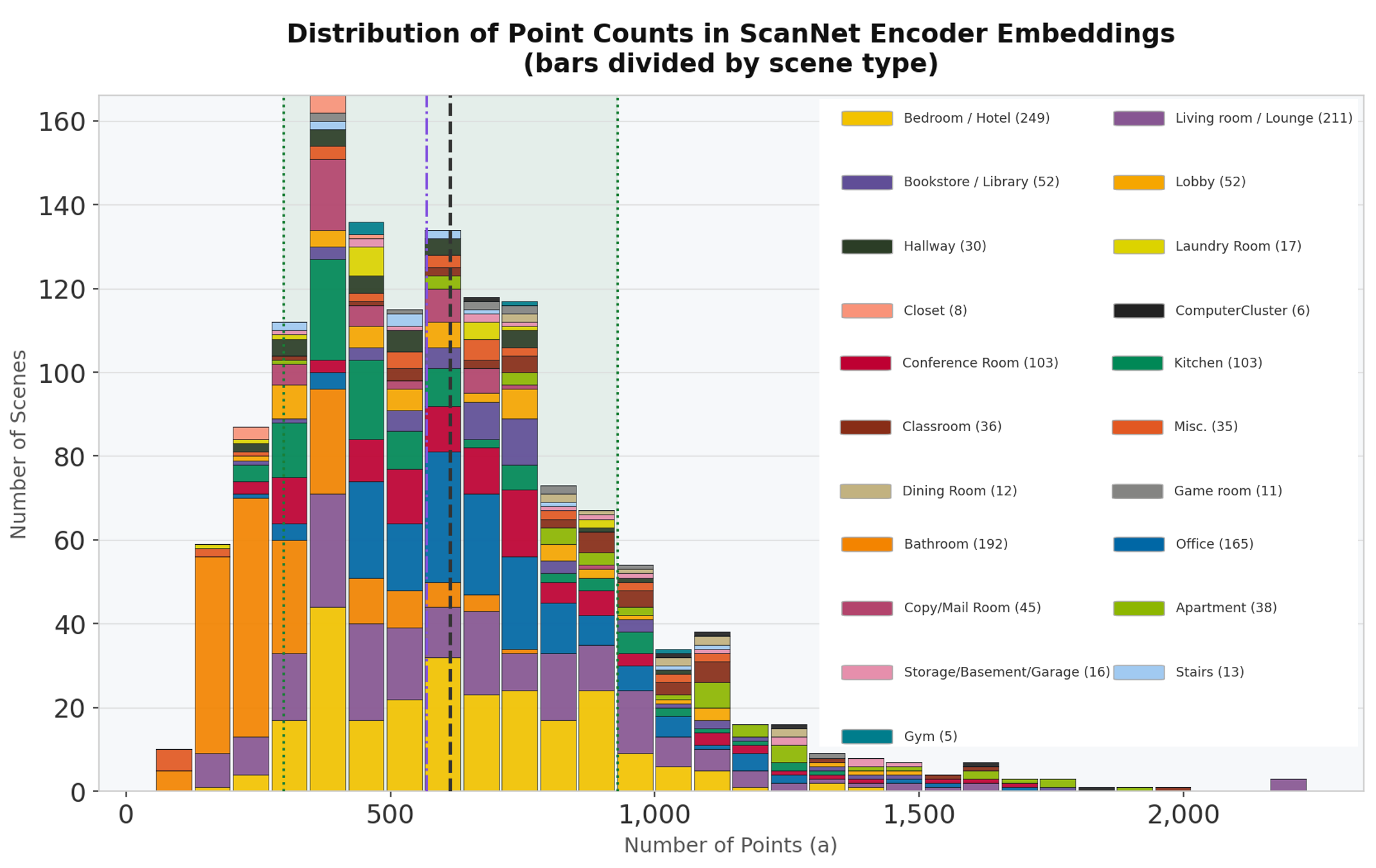}
\caption{Voxel count per ScanNet scene for a given label after voxelization. Wide variation motivates the voxel-level independence assumption for tractable MI estimation.}
\label{fig:point_counts}
\end{figure}

\paragraph{Point count distribution:} Fig.~\ref{fig:point_counts} shows point counts per scene after voxelization. 512-dimensional Sonata Encoder embeddings of Scannet scenes range from $\sim$10K to $>$300K voxels.

\begin{figure}[h]
\centering
\includegraphics[width=0.85\textwidth]{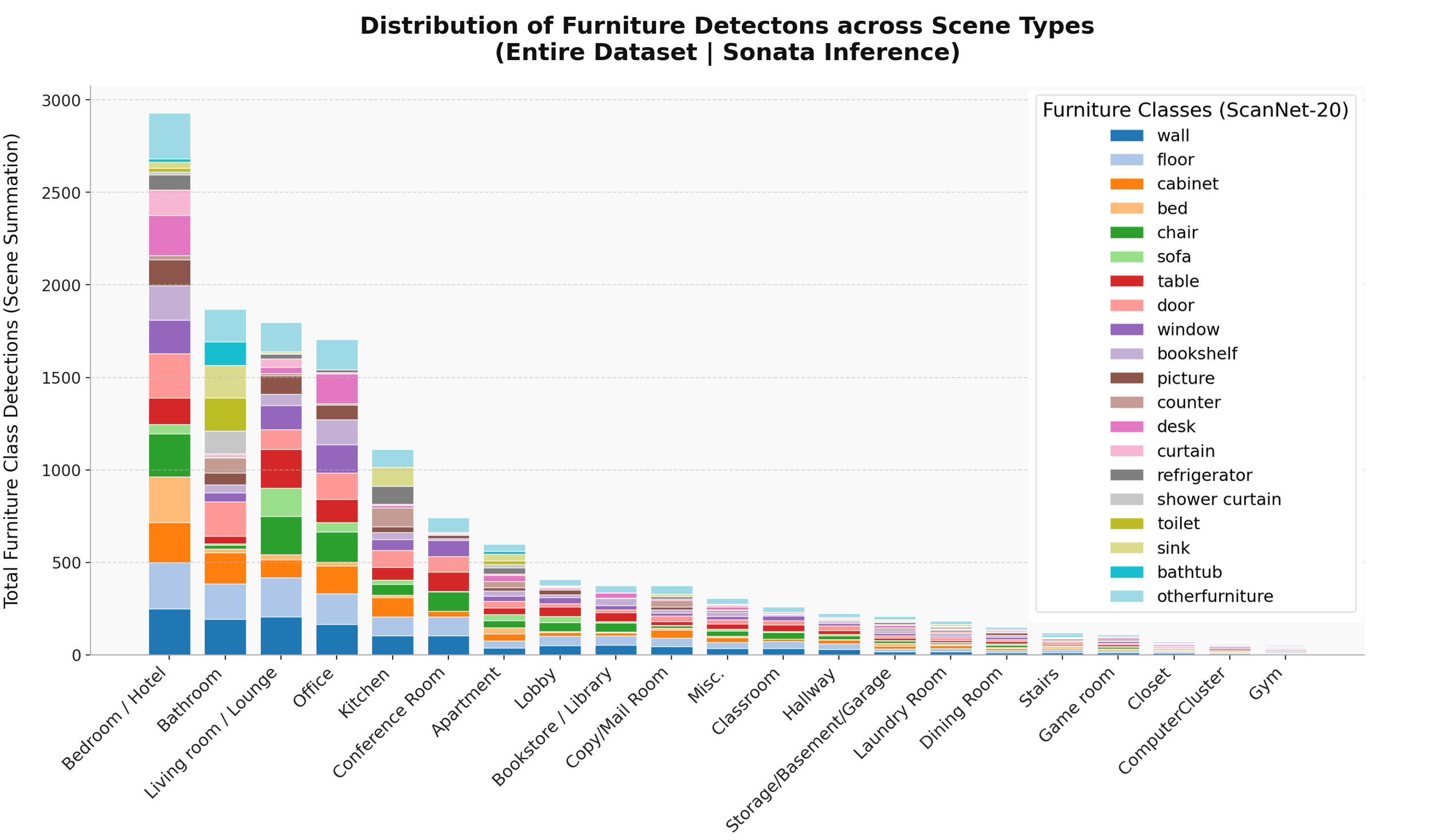}
\caption{Furniture distribution by room type. Strong furniture--room correlations make privacy-utility disentanglement challenging.}
\label{fig:furniture_hist}
\end{figure}

\paragraph{Furniture distribution by room type:} Fig.~\ref{fig:furniture_hist} reveals that certain objects correlate strongly with specific room types (e.g., beds in bedrooms, desks in offices), while others (walls, floors) are ubiquitous. This correlation makes the privacy-utility trade-off nontrivial: suppressing room-type information must not degrade furniture features that co-occur with specific rooms.

\begin{figure}[h]
\centering
\includegraphics[width=0.7\textwidth]{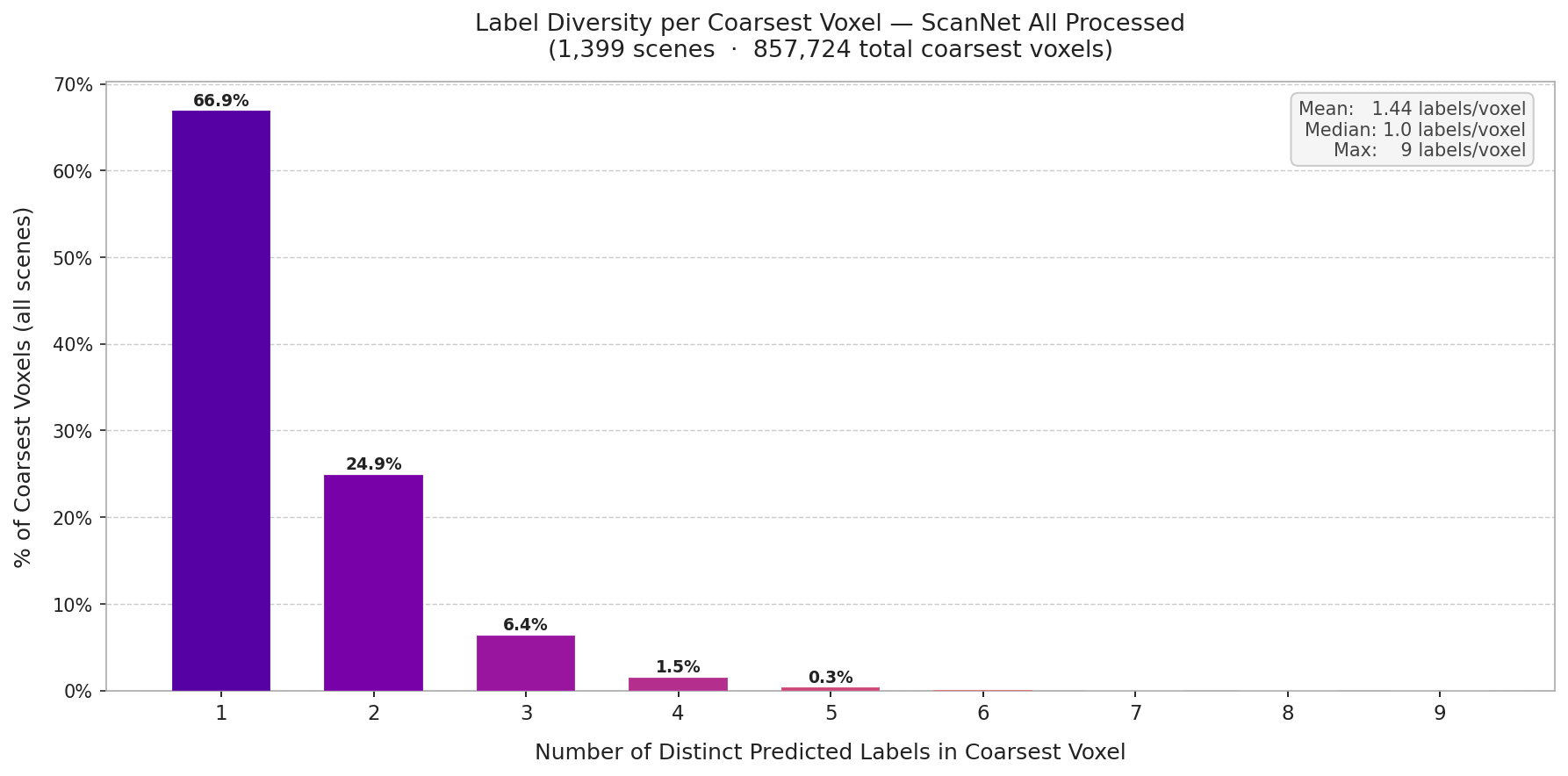}
\caption{Distinct furniture labels per voxel after Sonata encoding. Majority voting resolves multi-label voxels.}
\label{fig:voxel_diversity}
\end{figure}

\paragraph{Voxel label diversity:} During encoding, Sonata downsamples raw points into voxels that may aggregate multiple semantic labels. Fig.~\ref{fig:voxel_diversity} shows the number of distinct furniture labels per voxel. Most voxels are semantically pure (1 label), but boundary voxels contain 2--3+ labels. We assign each voxel the majority label among its constituent points.

\begin{figure}[h]
\centering
\includegraphics[width=0.85\textwidth]{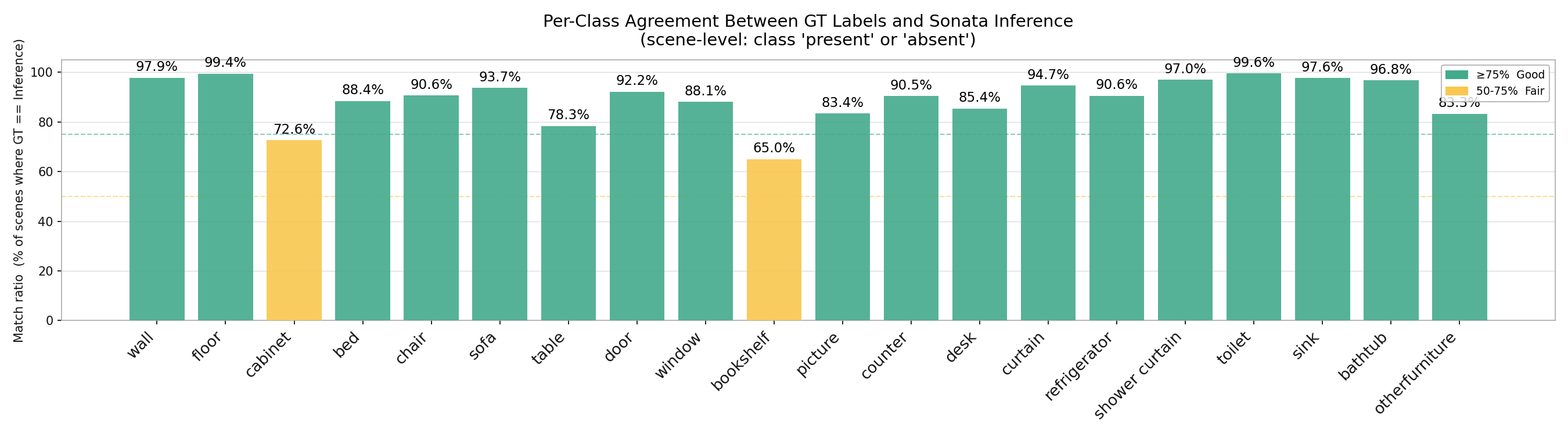}
\caption{Match ratio between Sonata's predicted segmentation and ScanNet ground-truth labels.}
\label{fig:inference_gt}
\end{figure}

\paragraph{Sonata inference vs.\ ground truth:} Fig.~\ref{fig:inference_gt} compares the Sonata encoder's segmentation against ScanNet ground-truth annotations, providing a baseline for interpreting the utility MI values in the main paper.

\section{Architecture and Training}
\label{app:architecture}

\paragraph{Sonata encoder (frozen):} We use the PTv3-based Sonata model~\cite{sonata} with 46.1M parameters, pre-trained via self-distillation on 140K point clouds from ScanNet, Structured3D, and S3DIS. The encoder produces a 512-dimensional embedding per voxel and is frozen throughout {\tt RoboShape} training.

\paragraph{Projection layer $\phi_\Theta$ (trainable):} A feedforward network: \texttt{Linear(512, 256)} $\to$ \texttt{ReLU} $\to$ \texttt{Linear(256, 128)} $\to$ \texttt{ReLU} $\to$ \texttt{Linear(128, 64)}, totaling 172,480 trainable parameters. No batch normalization, dropout, or output normalization is applied.

\paragraph{MI estimator networks $\{F_{\omega_i}\}_{i=1}^{3}$:} Each of the three DV-based estimators has two hidden layers (64 and 32 nodes, ReLU) and a scalar output. Input dimensions depend on the MI term: 576 for $I(\mathbf{z}; \mathbf{x})$, 65 for $I(\mathbf{z}; L(\mathbf{x}))$ and $I(\mathbf{z}; S(\mathbf{x}))$. These networks are discarded after training.

\paragraph{Training:} We train for 20 epochs using Adam with learning rate $10^{-3}$ for $\phi_\Theta$ and $10^{-4}$ for the MI estimators. The objective weights are $\gamma = 0$, $\lambda = 1$, with the privacy weight $\mu$ selected per dataset via the Pareto frontier between utility MI and privacy MI.

\section{Isaac Gym Simulation}
\label{app:isaacgym}

We train PPO policies in NVIDIA Isaac Gym~\cite{makoviychuk2021isaac} to test whether MI-based privacy shaping transfers to closed-loop robotic control. Table~\ref{tab:isaacgym} summarizes the configuration.

\begin{table}[h]
\centering
\caption{\textbf{Isaac Gym PPO configuration.}}
\label{tab:isaacgym}
\begin{tabular}{ll}
\toprule
\textbf{Parameter} & \textbf{Value} \\
\midrule
Environments & 52 (40 train, 12 test; balanced bedroom / non-bedroom) \\
Agent & Point-mass \\
Observation & Per-voxel embeddings ($\mathbb{R}^{512}$ or $\mathbb{R}^{64}$) \\
Action & 2D velocity $(v_x, v_y)$ \\
Policy network & Actor-Critic: 256 $\to$ 128 $\to$ 2 \\
Learning rate & $3 \times 10^{-4}$ \\
Discount $\gamma_{\mathrm{PPO}}$ & 0.99 \\
Clip ratio & $[0.8, 1.2]$ \\
PPO updates per rollout & 4 \\
Total epochs & 50 $\times$ 200 steps \\
\bottomrule
\end{tabular}
\end{table}

\section{Codebase}
\label{app:codebase}

The full codebase is available at 
. Table~\ref{tab:codebase} summarizes the repository structure.

\begin{table}[h]
\centering
\small
\caption{\textbf{Repository structure.}}
\label{tab:codebase}
\begin{tabular}{p{4.5cm}p{8cm}}
\toprule
\textbf{File / Directory} & \textbf{Description} \\
\midrule
\texttt{roboshape.py} & Model definition \\
\texttt{train.py} & Main training loop (dual MI optimization) \\
\texttt{test.py} & Inference and testing \\
\texttt{roboshape\_inference.py} & Loads trained $\phi_\Theta$, compresses embeddings \\
\midrule
\texttt{data\_prep/} & Preprocessing (ScanNet, Matterport3D, ARKitScenes) \\
\texttt{classifiers/} & Baseline evaluation (noisy, random encoder) \\
\texttt{roboshape\_isaacgym/} & PPO training and ScanNet navigation environment \\
\texttt{src/} & Core architecture and data utilities \\
\texttt{plots/} & Visualizations \\
\texttt{enviroment.yaml} & Conda environment (PyTorch, CUDA) \\
\bottomrule
\end{tabular}
\end{table}

\section{Broader Impact}
\label{app:impact}

{\tt RoboShape} is designed to enable privacy-aware sharing of 3D scene representations in multi-robot systems and cloud robotics pipelines. The primary beneficiaries are occupants of spaces scanned by robots, who gain information-theoretic guarantees that room-function information is suppressed before representations leave the device. The MI-based framework reduces mutual information between embeddings and room-type labels but does not constitute a formal privacy guarantee in the sense of differential privacy; an adversary with auxiliary information could potentially recover partial room-type information. We view {\tt RoboShape} as a practical defense layer and encourage responsible deployment.

\end{document}